\documentclass{INTERSPEECH2023}

\interspeechcameraready 

\usepackage{color}
\usepackage{multirow}
\usepackage[table,xcdraw]{xcolor}
\usepackage{amsmath}
\usepackage{subcaption} 
\DeclareMathOperator*{\argmax}{arg\,max}

\newcommand{\proposal}{LanSER}

\title{LanSER: Language-Model Supported Speech Emotion Recognition}
\name{Taesik Gong$^1$, Josh Belanich$^{2*}$, Krishna Somandepalli$^{2*}$, Arsha Nagrani$^2$, Brian Eoff$^2$,  Brendan Jou$^2$}
\address{
  $^1$KAIST, Republic of Korea\\
  $^2$Google Research}
\email{taesik.gong@kaist.ac.kr, \{joshbelanich,ksoman,anagrani,beoff,bjou\}@google.com}

\usepackage{filecontents} 

\makeatletter
\def\bstctlcite{\@ifnextchar[{\@bstctlcite}{\@bstctlcite[@auxout]}}
\def\@bstctlcite[#1]#2{\@bsphack
  \@for\@citeb:=#2\do{%
    \edef\@citeb{\expandafter\@firstofone\@citeb}%
    \if@filesw\immediate\write\csname #1\endcsname{\string\citation{\@citeb}}\fi}%
  \@esphack}
\makeatother

\begin{document}
\bstctlcite{IEEEexample:BSTcontrol}

\maketitle

\def\thefootnote{*}\footnotetext{Equal contribution.}\def\thefootnote{\arabic{footnote}}

\begin{abstract}
Speech emotion recognition (SER) models typically rely on costly human-labeled data for training, making scaling methods to large speech datasets and nuanced emotion taxonomies difficult. 
We present \proposal{}, a method that enables the use of unlabeled data by inferring weak emotion labels via pre-trained large language models through weakly-supervised learning. 
For inferring weak labels constrained to a taxonomy, we use a textual entailment approach that selects an emotion label with the highest entailment score for a speech transcript extracted via automatic speech recognition. Our experimental results show that models pre-trained on large datasets with this weak supervision outperform other baseline models on standard SER datasets when fine-tuned, and show improved label efficiency. Despite being pre-trained on labels derived only from text, we show that the resulting representations appear to model the prosodic content of speech.
\end{abstract}

\noindent\textbf{Index Terms}: speech emotion recognition, large language models, weakly-supervised learning

\section{Introduction}

In conversations, humans rely on both \textit{what is said} (i.e., lexical content), and \textit{how it is said} (i.e., prosody), to infer the emotion expressed by a speaker.
State-of-the-art methods in speech emotion recognition (SER) leverage the interplay of these two components for modeling emotional expression in speech.
However, such methods still show limitations on in-the-wild scenarios due to the variability in natural speech, and the reliance on human ratings using limited emotion taxonomies.
Extending model training to large, natural speech datasets labeled by humans for nuanced emotion taxonomies is expensive and is further complicated by the subjective nature of emotion perception.

Despite both lexical content and prosody being complementary for emotion perception, the two components are \textit{correlated}, and in many cases the content is predictive of the prosody.
For example, when someone says, ``I won the lottery" – an upbeat and lively prosody would sound congruent, and one might perceive the emotional expression as elation or triumphant.
In this work, we investigate how we might leverage the emotions congruent with lexical content in large unlabeled speech datasets to serve as weak supervision for developing SER models.

We turn to Large Language Models (LLMs) to infer expressed emotion categories in textual content. Due to the knowledge they embed from pre-training on large text corpora~\cite{clip, gpt3}, LLMs have demonstrated capabilities in numerous downstream tasks~\cite{liang2022holistic}, including a few subjective tasks such as social and emotion reasoning~\cite{sap-etal-2019-social}.
In domains such as computer vision, LLMs were explored to reduce the need for labeled data, e.g., for visual question answering~\cite{yang2021just}.
However, to our knowledge, they have not been studied for emotion recognition tasks, particularly from natural speech.

\begin{figure}[t]
\centering
\includegraphics[width=1\linewidth]{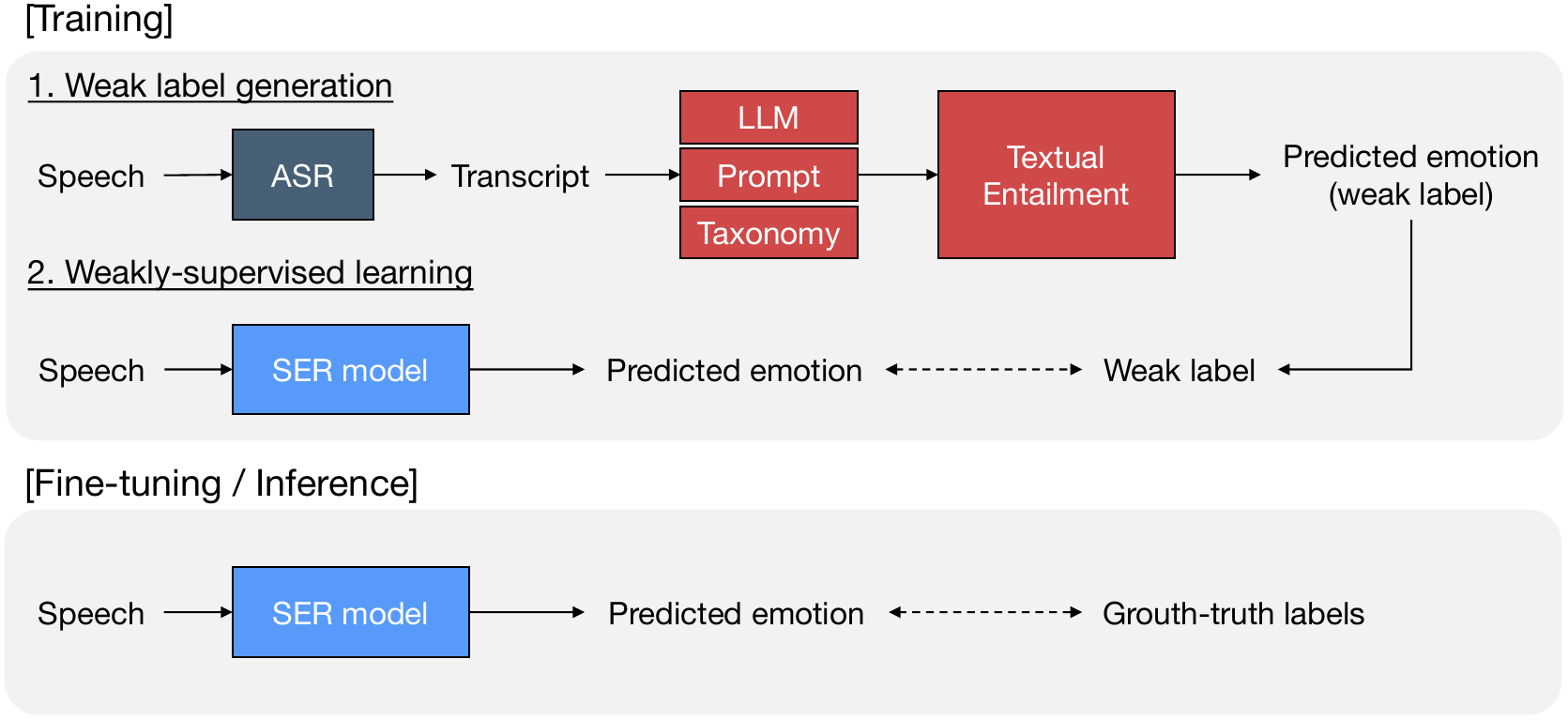}
\caption{The overview of \proposal{}. LLMs and textual entailment are used to infer weak emotion labels from speech content which are used to pre-train a SER model. 
}
\label{fig:overview}
\vspace{-0.6cm}
\end{figure}

We propose \textit{\proposal{}}, that uses LLMs to infer emotion categories from speech content i.e., transcribed text, which serve as weak labels for SER (Figure~\ref{fig:overview}).
Overall, \proposal{} enables pre-training a SER model on large speech datasets without human labels by (1) extracting text transcripts from utterances using automatic speech recognition (ASR), (2) using pre-trained LLMs to infer weak emotion labels with an engineered prompt and predetermined taxonomy, and (3) pre-training the SER model with the weak labels.
We demonstrate that \proposal{} improves SER performance and label efficiency by fine-tuning on benchmark datasets.
Moreover, we show that despite the emotion labels being derived from speech content only, \proposal{} captures speech prosody information that is relevant to SER.

\section{Related Work}

\noindent \textbf{SER with LLMs}: 
Recently, LLMs were used to generate pseudo-labels for semi-supervised learning for speech sentiment analysis~\cite{ser_llm}. Here, LLMs were fine-tuned on a \textit{labeled} sentiment dataset to explore narrow sentiment classes of negative, positive, and neutral. In contrast, our work avoids fine-tuning LLMs on task-specific datasets by inferring weak labels via textual entailment, enabling exploration with wider emotion taxonomies. In the context of multi-modal emotion recognition, MEmoBERT~\cite{memobert} used audio, visual, and text information with prompt learning for unsupervised emotion recognition. Herein, the visual model is pre-trained on a large labeled emotion dataset. In contrast, in our work, pre-training on large human-annotated emotion datasets is not necessary.


\noindent \textbf{Self-supervised learning}: Self-supervised learning has become a popular method using large amounts of unlabeled speech data for pre-training~\cite{wav2vec2, cap}. Recent studies found that large pre-trained models via self-supervised learning show effectiveness in various downstream speech tasks~\cite{yang2021superb}, including many paralinguistic tasks~\cite{cap}. We view self-supervised learning and our weak supervision from LLMs as complementary, since the two methodologies can be combined for training SER models.

\section{Methodology}


An overview of the training and inference process of \proposal{} in shown in Figure~\ref{fig:overview}. During pre-training, we use ASR to generate transcripts from speech utterances, which are fed into a LLM with appropriate prompt to extract weak emotion labels in predetermined taxonomy via textual entailment. These labels are used to pre-train a SER model via weakly-supervised learning. The pre-trained SER model can then either be used directly to output emotion predictions according to the emotion taxonomy used to extract weak labels, or can be adapted for a different taxonomy or dataset by fine-tuning.

We note that the emotions inferred using LLMs from speech content are proxies for the emotion being expressed, and may not capture the larger context or intent of the speaker. Thus, we treat them as ``weak'' emotion labels in our work.

\subsection{Weak label generation via textual entailment}

\begin{figure}[t]
\centering
\scalebox{0.95}{\includegraphics[width=1\linewidth]{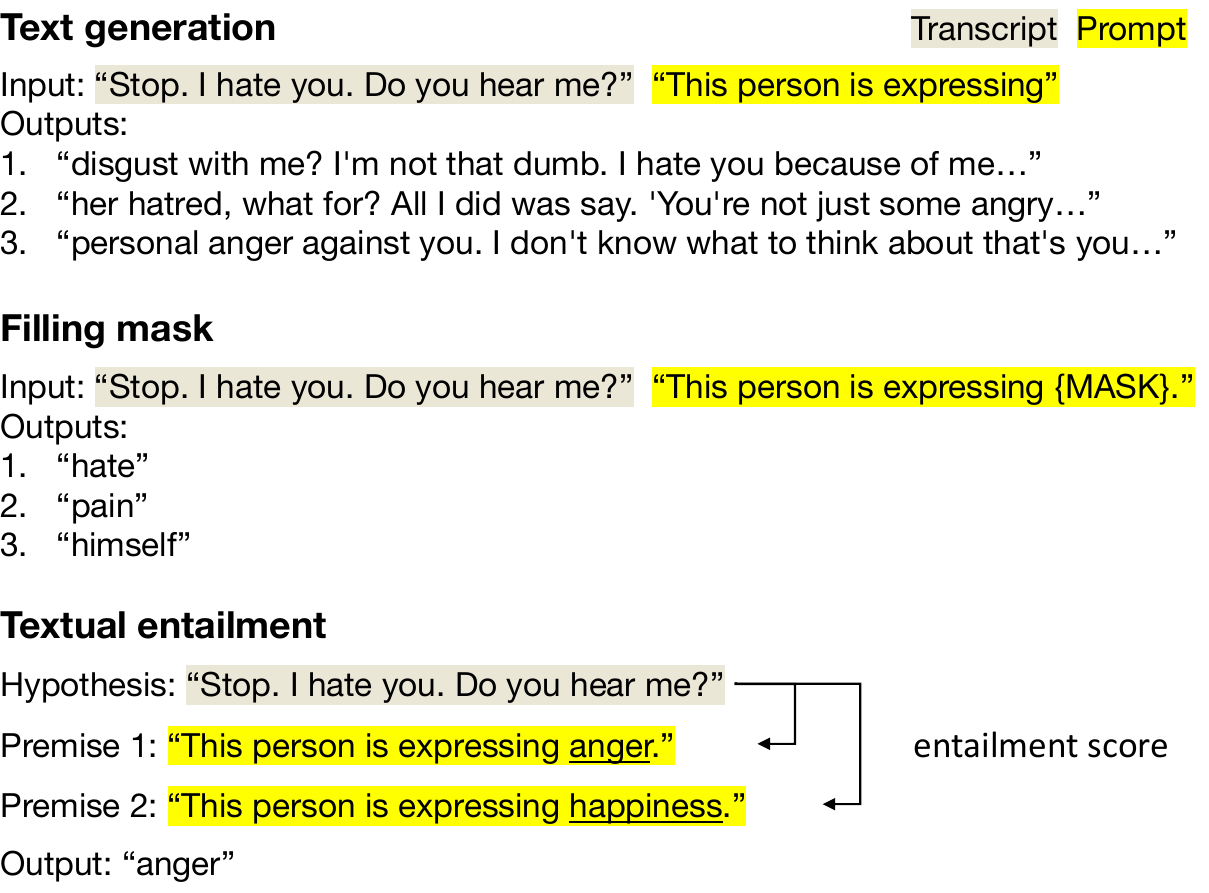}}
\vspace{-0.2cm}
\caption{Comparison of three weak label generation approaches: text generation, filling mask, and textual entailment.}
\vspace{-0.3cm}
\label{fig:llm}
\end{figure}

There are multiple ways to use LLMs for extracting weak emotion labels. Two dominant approaches in the literature are (i) text generation~\cite{gpt3} and (ii) filling mask~\cite{bert, clip, memobert}. Figure~\ref{fig:llm} demonstrates the behaviors of text generation and filling mask for weak emotion label prediction. We used representative LLMs for each approach: GPT-2 for text generation and BERT~\cite{bert} for filling mask. While these approaches show some success, the common limitation in a zero-shot setting is that they often output undesirable ``noise'', like irrelevant words (text generation), or non-emotional responses (e.g., ``himself'' in filling mask in the Fig.~\ref{fig:llm}). 

Thus, we want to constrain the LLM model to output only words relevant to emotion perception. To this end, we use textual entailment~\cite{zs_nli} to generate weak labels that also allows us to constrain the emotion taxonomy apriori. Figure~\ref{fig:llm} illustrates the entailment-based weak emotion label generation; at a high-level, this method calculates the entailment scores between an input transcript (called \textit{hypothesis}) and prompts with candidate labels from the taxonomy (called \textit{premise}), and then selects the item with the highest score as the weak label. Formally, let $x \in \mathcal{X}$ denote ASR transcripts from speech and $y \in \mathcal{Y}$ denote a candidate label in taxonomy $\mathcal{Y}$. A prompting function $g(\cdot)$ prepends a predefined prompt to the given input. $f(x, g(y))$ denotes the entailment score between a hypothesis $x$ and a prompted label $g(y)$. The resulting weak emotion label $\hat{y}$ for a given transcript $x$ is calculated as:
\begin{equation}
    \hat{y}:= \argmax_{y\in\mathcal{Y}} f(x, g(y)).
\end{equation}
The entailment scoring function $f$ is a function typically parameterized by a neural network and fine-tuned on the entailment task. In our case, we use RoBERTa~\cite{roberta} fine-tuned on the Multi-genre Natural Language Inference (MNLI)~\cite{mnli} dataset. The MNLI dataset is composed of hypothesis and premise pairs for diverse genres, which is specialized for the textual entailment approach, and do not explicitly focus on emotion-related concepts.

\subsection{Prompt engineering}
Prompt engineering is a task-specific description embedded in inputs to LLMs (e.g., a question format)~\cite{10.1145/3560815}. It is a critical component affecting zero-shot performance of LLMs on various downstream tasks~\cite{clip, ier_prompt, prompt}. In Section~\ref{sec:prompt} we explore various prompts in order to understand the impact of prompt engineering for the entailment task. Ultimately, we found that the prompt ``The emotion of the conversation is \{\}.'' performed best, and we use this prompt throughout our experiments.

\subsection{Taxonomy}\label{sec:tax}

The choice of emotion taxonomy is critical in developing SER models as emotion perception and expression is nuanced. Common SER benchmarks typically use 4--6 emotion categories~\cite{iemocap,crema_d}, which do not capture the variability in emotion perception~\cite{cowen2017self}. Thus we experiment with BRAVE-43, a finer-grained taxonomy~\cite{brave}. We adopted and modified the BRAVE taxonomy which originally contains 42 self-reported emotions labels. We converted several two-word emotions to one-word emotions for simplicity and added ``shock'' to capture a negative version of ``surprise'', resulting in a total of 43 categories. Note this taxonomy is not speech-specific. We investigate the impact of taxonomy selection in Section~\ref{sec:taxonomy}. We expect fine-grained taxonomies to help learn effective representations by using the high degree of the expressiveness of LLMs. 
\section{Experiments}
Our overarching hypothesis is that, given a sufficiently large amount of data, pre-training speech-only models on weak emotion labels derived from text improves performance on SER tasks. As such, throughout this paper, we focus on speech-only emotion recognition models. Additionally, our goal is not to obtain state-of-the-art results on downstream tasks but to assess, given a fixed model capacity, whether models pre-trained via \proposal{} achieve improved performance.
\subsection{Data preparation} 

\noindent\textbf{Pre-training data}: We investigate two large-scale speech datasets for \proposal{} pre-training: People's Speech~\cite{peoples_speech} and Condensed Movies~\cite{condensed_movies}. People's Speech is currently the largest English speech recognition corpus, containing 30K hours of general speech. Condensed Movies is comprised of 1,000 hours of video clips from 3,000 movies, where we use only the audio. We explore these two large-scale speech datasets to understand the impact of the amount of data and their distributions; while People's Speech has more samples from less emotional data sources (e.g., government, interview, health, etc.), Condensed Movies has fewer samples from a more emotional data source (movies). We use Whisper ASR~\cite{whisper} (``small'' variant) to segment and generate transcripts for People's Speech and Condensed Movies datasets, resulting in 4,321,002 and 1,030,711 utterances, respectively. 

\noindent\textbf{Downstream tasks}: 
We use two common SER benchmarks for downstream tasks: IEMOCAP~\cite{iemocap} and CREMA-D~\cite{crema_d}. IEMOCAP is an acted, multi-speaker database containing 5,531 audio clips from 12 hours of speech. We follow the commonly used four-class (anger, happiness, sadness, and neutral) setup~\cite{memobert, zhao-etal-2021-missing, yang2021superb, cap} and use speaker-independent train:val:test splits. CREMA-D has 7,441 audio clips collected from 91 actors. An important characteristic of CREMA-D is that it is linguistically constrained, having only 12 sentences each presented using six different emotions (anger, disgust, fear, happy, neutral, and sad). We use CREMA-D to validate that our models indeed learn prosodic representations, and do not just learn to use language to predict the emotional expression.

\subsubsection{Baselines}
We compare \proposal{} models fine-tuned on downstream datasets with the following four baselines:

    \noindent \textbf{Majority}: Output the most prevalent class in the dataset~\cite{zs_nli}. 
    
    \noindent \textbf{GT Transcript + Word2Vec}~\cite{w2v}: Each word in a ground-truth transcript is converted to a Word2Vec embedding. We compute the cosine similarity between the averaged transcript embedding and each class label, outputting the class with the highest similarity.
    
    \noindent \textbf{GT Transcript + LLM + Entailment}~\cite{zs_nli}: Using the same methodology for producing weak labels, we process the ground-truth transcript with an LLM and entailment to output a classification according to the dataset's taxonomy.
    
    \noindent \textbf{Supervised}: Supervised learning using the same model architecture as \proposal{} but without pre-training.

We include two language-based methods (Word2Vec and Entailment) to better understand how \proposal{} compares with models using lexical content alone. Note that the language baselines assume GT transcripts are available. In practice, these baselines would require an ASR pipeline to get transcripts, which may involve additional computational and developmental cost.

\subsubsection{Implementation}
We extracted mel-spectrogram features (frame length 32ms, frame steps 25ms, 50 bins from 60--3600Hz) from the audio waveforms as input to the model and used ResNet-50~\cite{resnet} as the backbone network for training. For both pre-training and fine-tuning, we minimized the cross-entropy loss with the Adam~\cite{adam} optimizer and implemented in TensorFlow~\cite{tensorflow}.

For pre-training, we adopted a warm-up learning rate schedule where the rate warmed up for the initial 5\% of updates to a peak of $5\times10^{-4}$ and then linearly decayed to zero. We used a batch size of 256 and trained for 100K iterations.

For fine-tuning on the downstream tasks, we loaded the pre-trained weights and used a fixed learning rate of $10^{-4}$. We set the batch size as 64 and trained for 10K iterations. We split the downstream datasets into a 6:2:2 (train:valid:test) ratio, and selected the best model on the validation set for testing.

\begin{table}[t]
\centering
\caption{Accuracy of extracted weak emotion labels with various prompts. \{\} indicates the masked position.}
\vspace{-0.2cm}
\resizebox{0.9\linewidth}{!}{%
\begin{tabular}{lr}
\hline
\multicolumn{1}{c}{\textbf{Prompts}} & \multicolumn{1}{c}{\textbf{Acc.}} \\ \hline
This example is \{\}. & 42.0\% \\
I am \{\}.~\cite{memobert} & 39.9\% \\
I feel \{\}. & 41.8\% \\
I am feeling \{\}. & 45.0\% \\
This person is expressing \{\} emotion. & 43.7\% \\
A speech seems to express a feeling like \{\}.~\cite{ier_prompt} & 38.0\% \\
A transcript seems to express a feeling like \{\}.~\cite{ier_prompt} & 38.9\% \\
A conversation seems to express some feelings like \{\}.~\cite{ier_prompt} & 39.0\% \\
The emotion of the conversation is \{\}. & \textbf{45.6\%} \\
The emotion of the previous conversation is \{\}. & 44.1\% \\
The overall emotion of the conversation is \{\}. & 45.1\% \\ \hline
\end{tabular}%
}
\vspace{-0.3cm}
\label{tab:prompt}
\end{table}

\begin{table*}[t]
\centering
\caption{Unweighted accuracy (\%) of fine-tuning for downstream tasks with varying the percentage of fine-tuning data (10\%, 30\%, 50\%, 70\%, and 100\%). Bold fonts indicate the highest accuracy.}
\vspace{-0.2cm}
\resizebox{0.7\linewidth}{!}{\begin{tabular}{clrrrrr}
\hline
\textbf{Downstream task} & \multicolumn{1}{c}{\textbf{Method \ \textbackslash \ Fine-tuning data:}} & \multicolumn{1}{c}{\textbf{10\%}} & \multicolumn{1}{c}{\textbf{30\%}} & \multicolumn{1}{c}{\textbf{50\%}} & \multicolumn{1}{c}{\textbf{70\%}} & \multicolumn{1}{c}{\textbf{100\%}} \\ \hline
 & Majority & 30.9\% & 30.9\% & 30.9\% & 30.9\% & 30.9\% \\
 & GT Transcript + Word2Vec~\cite{w2v} & 34.9\% & 34.9\% & 34.9\% & 34.9\% & 34.9\% \\
 & GT Transcript + Entailment~\cite{zs_nli} & \cellcolor[HTML]{FFFFFF}47.5\% & \cellcolor[HTML]{FFFFFF}47.5\% & \cellcolor[HTML]{FFFFFF}47.5\% & \cellcolor[HTML]{FFFFFF}47.5\% & \cellcolor[HTML]{FFFFFF}47.5\% \\
 & Supervised & 38.5\% & 41.8\% & 45.5\% & 46.0\% & 47.6\% \\ \cline{2-7} 
& \proposal{} (People's Speech) & 42.0\% &	47.1\% & 47.2\% & \textbf{50.0}\%	& 50.6\% \\
\multirow{-6}{*}{IEMOCAP~\cite{iemocap}} & \proposal{} (Condensed Movies) & \textbf{50.0}\% & \textbf{51.7}\% & \textbf{48.0}\% & 45.4\% & \textbf{54.5}\% \\ \hline
 & Majority & 17.1\% & 17.1\% & 17.1\% & 17.1\% & 17.1\% \\
 & GT Transcript + Word2Vec~\cite{w2v} & 19.1\% & 19.1\% & 19.1\% & 19.1\% & 19.1\% \\
 & GT Transcript + Entailment~\cite{zs_nli} & 16.1\% & 16.1\% & 16.1\% & 16.1\% & 16.1\% \\
 & Supervised & 37.8\% & 43.2\% & 48.2\% & 53.4\% & 57.2\% \\ \cline{2-7} 
 & \proposal{} (People's Speech) & 35.5\% &	48.2\% & 51.5\%	& 52.7\% & 55.8\% \\ 
\multirow{-6}{*}{CREMA-D~\cite{crema_d}} & \proposal{} (Condensed Movies) & \textbf{43.7}\% & \textbf{49.9}\% & \textbf{52.2}\% & \textbf{53.6}\% & \textbf{58.7}\% \\\hline
\end{tabular}}
\vspace{-0.3cm}
\label{tab:fine_tuning}
\end{table*}

\begin{table}[t]
\centering
\caption{Unweighted accuracy (\%) for fine-tuning on downstream tasks. LanSER (random labels) is pre-trained on Condensed Movies with BRAVE taxonomy labels assigned randomly.}
\vspace{-0.2cm}

\resizebox{0.9\linewidth}{!}{%
\begin{tabular}{clr}
\hline
\textbf{Downstream task} & \multicolumn{1}{c}{\textbf{Method}} &
\multicolumn{1}{c}{\textbf{Accuracy}} \\ \hline
 & LanSER (random labels) & 47.6\% \\ 
\multirow{-2}{*}{IEMOCAP~\cite{iemocap}}  & \proposal{} (weak labels) & \textbf{54.5}\% \\ \hline
 & LanSER (random labels) & 50.6\% \\ 
 
\multirow{-2}{*}{CREMA-D~\cite{crema_d}}  & \proposal{} (weak labels) & \textbf{58.7}\% \\ \hline
\end{tabular}
}
\vspace{-0.3cm}
\label{tab:random_labels}
\end{table}

\begin{table}[t]
\centering
\caption{Zero-shot unweighted accuracy (\%) of SER models.}
\vspace{-0.2cm}

\resizebox{0.9\linewidth}{!}{%
\begin{tabular}{clr}
\hline
\textbf{Downstream task} & \multicolumn{1}{c}{\textbf{Method}} &
\multicolumn{1}{c}{\textbf{Accuracy}} \\ \hline
 & Scratch & 22.9\% \\ 
& \proposal{} (People's Speech) & 30.9\% \\ 
\multirow{-3}{*}{IEMOCAP~\cite{iemocap}}  & \proposal{} (Condensed Movies) & \textbf{34.3}\% \\ \hline
 & Scratch & 16.3\% \\ 
& \proposal{} (People's Speech) & 15.9\% \\ 
 
\multirow{-3}{*}{CREMA-D~\cite{crema_d}}  & \proposal{} (Condensed Movies) & \textbf{23.5}\% \\ \hline
\end{tabular}
}
\vspace{-0.3cm}
\label{tab:zero_shot}
\end{table}

\subsection{Prompt engineering}\label{sec:prompt}
We investigated the impact of various prompts to infer weak emotion labels using IEMOCAP. We chose IEMOCAP because it has transcripts and human-rated labels with majority agreement referred here as ``ground-truth''. To evaluate the prompts, we compute accuracy by comparing the weak labels with the ground-truth. We also examined prompts used in previous emotion recognition studies~\cite{ier_prompt, memobert} and modified a few vision-specific prompts~\cite{ier_prompt} for our study by replacing words such as ``photo'' or ``image'' with ``speech''.

Table~\ref{tab:prompt} shows the accuracy for each prompt. 
The prompt (``I am \{\}.)'' used in the related sentiment work~\cite{memobert} was not as effective at capturing emotional signals. Similarly, adapting vision-specific prompts~\cite{ier_prompt} was ineffective. This suggests that it is worthwhile to tailor the prompt to the SER task. Among the prompts we explored, ``The emotion of the conversation is \{\}.'' had the highest accuracy. We adopt this prompt to infer weak labels in all our experiments. We leave additional prompt tuning~\cite{prompt_tuning} as future work.

\subsection{Fine-tuning}

We fine-tune all models on the downstream tasks to evaluate their label efficiency and performance. To measure label efficiency, we varied the percentage of seen training data from 10\% to 100\% for each dataset. Table~\ref{tab:fine_tuning} shows the result. ``\proposal{} (People's Speech)'' means pre-training with Peoples Speech, while ``\proposal{} (Condensed Movies)'' refers to pre-training with Condensed Movies. In all cases, we used the BRAVE taxonomy (see Sec.~\ref{sec:tax}) as the label space.


First,  NLP baselines (Word2Vec and Entailment) fail on CREMA-D, as they only use lexical speech content. Interestingly, \proposal{}'s results on CREMA-D suggest that the model can learn prosodic representations via weak supervision from LLMs. We attribute this result to  pre-training with large-scale data, and it offers evidence to our hypothesis that speech and text emotions are correlated enough that SER models can learn to use prosodic features even with labels from text only given a sufficiently large amount of data.

Overall, \proposal{} outperforms the NLP and majority class baselines. Notably, \proposal{} pre-trained with the Condensed Movies showed improved accuracy than with the People's Speech. While People's Speech is comprised of fairly neutral speech data (e.g., government, interviews, etc.), Condensed Movies is comprised of movies having more expressive speech; from the emotion recognition perspective, Peoples Speech might introduce more noise than Condensed Movies.

To assess that performance improvements are being driven by the emotion labels inferred using LLMs, and not just the scale of the pre-training data, we compare the fine-tuning performance of \proposal{} to a model pre-trained on Condensed Movies using random uniformly sampled labels. As shown in Table~\ref{tab:random_labels}, models pre-trained with weak labels outperform ones trained with random labels suggesting that the weak emotion labels inferred using LLMs are meaningful.


\subsection{Zero-shot classification accuracy}\label{sec:zero_shot}
A unique advantage of \proposal{} over self-supervised learning~\cite{wav2vec2, cap} is that it enables SER models to support zero-shot classification.
Table~\ref{tab:zero_shot} shows the zero-shot classification accuracy: for \proposal{}, SER models were pre-trained with the taxonomy of the downstream dataset instead of BRAVE and evaluated in a zero-shot setting. We use models with randomly initialized weights and no training as a lower-bound of performance, referred to as ``Scratch''. Overall, \proposal{} shows higher accuracy than the baseline, although not as good as fine-tuning. These results suggest the potential of training large SER models that can perform well on various downstream tasks, without further fine-tuning. Improving zero-shot performance further using our proposed framework is part of our future work.

\begin{figure}[t]
\captionsetup[subfigure]{justification=centering} 
    \centering
    \begin{subfigure}[t]{1\linewidth}
        \centering
        \begin{subfigure}[t]{0.49\linewidth}
            \centering
            \includegraphics[width=1\linewidth]{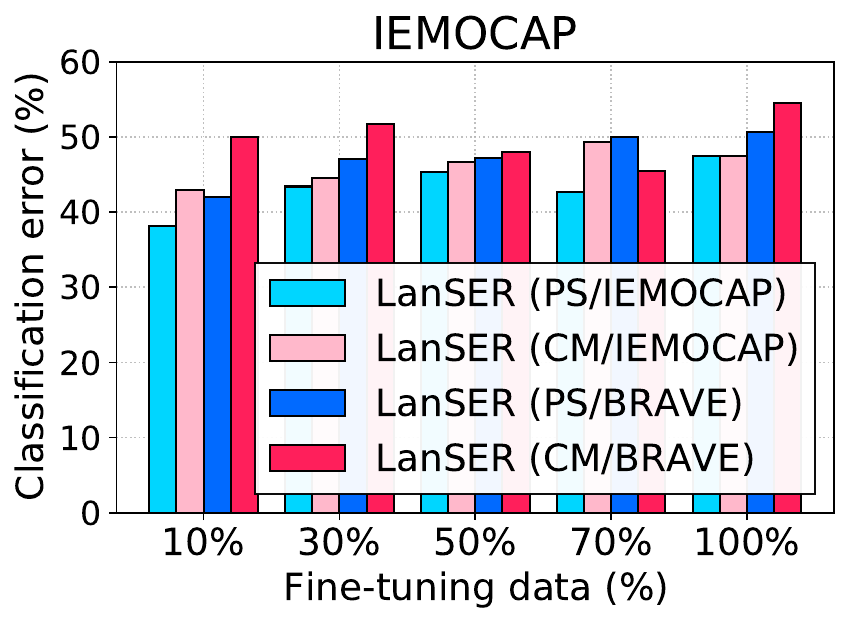}
            \label{fig:taxonomy:iemocap}
        \end{subfigure}
        ~
        \begin{subfigure}[t]{0.49\linewidth}
            \centering
            \includegraphics[width=1\linewidth]{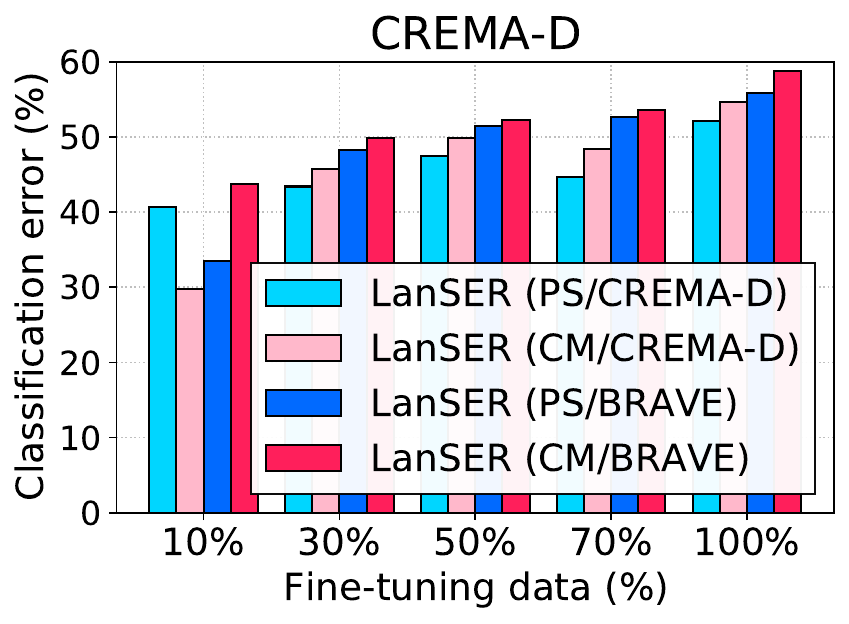}
            \label{fig:taxonomy:crema_d}
        \end{subfigure}
    \centering
    \end{subfigure}
    \vspace{-0.6cm}
    \caption{Impact of taxonomy selection for pre-training.}
    \vspace{-0.3cm}
    \label{fig:taxonomy}
\end{figure}

\subsection{Impact of taxonomy}\label{sec:taxonomy}

Figure~\ref{fig:taxonomy} shows the impact of taxonomy selection. We compared the BRAVE taxonomy with downstream task's taxonomies. ``PS'' and ``CM'' refers to People's Speech and Condensed Movies, respectively. ``IEMOCAP'', ``CREMA-D'', and ``BRAVE'' means taxonomy used to generate weak labels. As shown, pre-training with the finer taxonomy (BRAVE) shows generally better accuracy when fine-tuned, with 4.2\% accuracy improvement on average. This indicates that a fine-grained taxonomy is beneficial to learn effective representations by leveraging the high degree of the expressiveness of LLMs.
\section{Caveats}

Developing machine perception models of apparent emotional expression remains an open area of investigation. 
The models in this work do not aim to infer the internal emotional state of individuals, but rather model proxies from speech utterances. This is especially true when training on the output of LLMs, since LLMs may not take into account prosody, cultural background, situational or social context, personal history, and other cues relevant to human emotion perception. ASR transcription errors add another layer of noise.

The benchmark datasets we use in this work are relatively small and are labeled with limited emotion taxonomies. CREMA-D, while useful for its fixed lexical content, is an acted dataset where the  emotional expression of its utterances may not well-represent natural speech.


\section{Conclusion and Future Work}

In this work, we proposed \proposal{}, a novel language-model supported speech emotion recognition method that leverages large unlabeled speech datasets by generating weak labels via textual entailment using LLMs. Our experimental results showed that \proposal{} can learn effective emotional representations including prosodic features. 

We note several possible areas of future work. It may be possible to reduce the weak label noise via filtering mechanisms, or by modifying prompts to include more conversational context, like the previous and next utterances, or scene descriptions. Additionally, using LLMs to generate weak labels in an open-set taxonomy may better leverage their expressiveness. Finally, while in this work we used ResNet-50 as our backbone model, higher capacity models like Conformers~\cite{cap} might better capture the complex relationship between speech and emotion on the pre-training datasets we explored. We believe that the initial investigation and findings of this work provide valuable insights for future SER research on large-scale unlabeled data.

\clearpage
\bibliographystyle{IEEEtran}
\bibliography{ref.bib, \jobname}{}
\end{document}